\documentclass{article} 
\usepackage{iclr2025_delta,times}
\usepackage{graphicx}
\usepackage{enumitem}
\usepackage{hyperref}

\usepackage{amsmath,amsfonts,bm}

\def\eqref#1{equation~\ref{#1}}

\def\1{\bm{1}}

\DeclareMathAlphabet{\mathsfit}{\encodingdefault}{\sfdefault}{m}{sl}
\SetMathAlphabet{\mathsfit}{bold}{\encodingdefault}{\sfdefault}{bx}{n}

\usepackage{hyperref}
\usepackage{url}

\title{Image-Alchemy: Advancing Subject Fidelity in Personalized Text-to-Image Generation}

\author{Amritanshu Tiwari, Cherish Puniani, Kaustubh Sharma, Ojasva Nema \thanks{Equal contribution} \\
Indian Institute of Technology Roorkee, India \\
\texttt{\{amritanshu\_t@mfs, cherish\_p@me, kaustubh\_s@ee, ojasva\_n@mt\}.iitr.ac.in} \\
}

\iclrfinalcopy
\begin{document}

\maketitle

\begin{abstract}
Recent advances in text-to-image diffusion models, particularly Stable Diffusion, have enabled the generation of highly detailed and semantically rich images. However, personalizing these models to represent novel subjects based on a few reference images remains challenging. This often leads to catastrophic forgetting, over-fitting, or large computational overhead. We propose a two-stage pipeline that addresses these limitations by leveraging LoRA-based fine-tuning on the attention weights within the U-net of Stable Diffusion XL model. Next, we exploit the unmodified SDXL to generate a generic scene, replacing the subject with its class label. Then we selectively insert the personalized subject through a segmentation-driven Img2Img pipeline that uses the trained LoRA weights. The framework isolates the subject encoding from the overall composition, thus preserving SDXL’s broader generative capabilities while integrating the new subject in a high-fidelity manner. Our method achieves a DINO similarity score of 0.789 on SDXL, outperforming existing personalized text-to-image approaches.
\end{abstract}

\begin{figure}[h]
\begin{center}
    \includegraphics[width=1\linewidth]{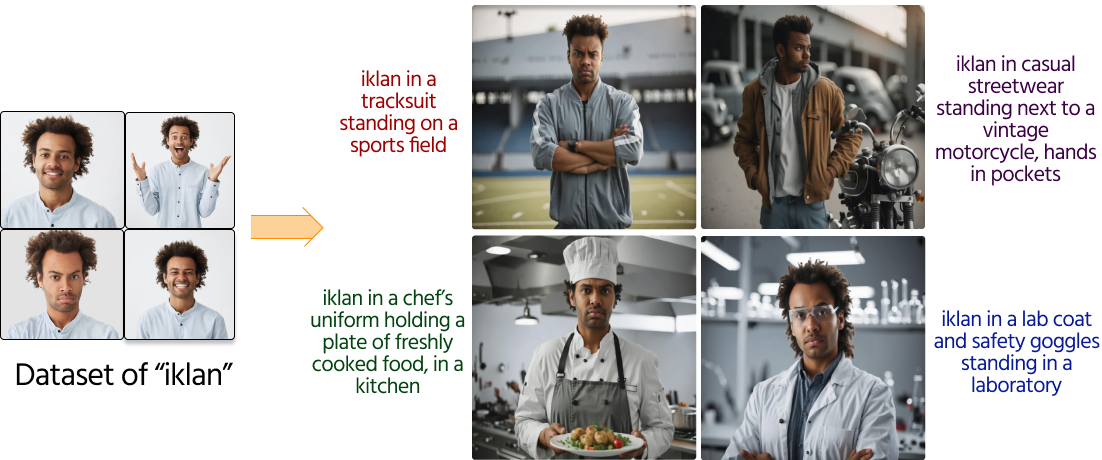} 
\end{center}
\caption{Image-Alchemy : Illustration of personalized generation for the subject “iklan”, the token chosen for the given person. The left panel shows four reference images used to fine-tune our model, and the right panel demonstrates four diverse outputs, highlighting how the learned subject adapts seamlessly to various prompts.}
\end{figure}

\vspace{-0.2 cm}
\section{Introduction}

Deep generative models (DGMs), particularly latent diffusion models have played a significant role in revolutionizing high resolution image synthesis. The state-of-the-art latent diffusion based image generation models, including Stable Diffusion(\cite{rombach2022highresolutionimagesynthesislatent}), Dall-e (\cite{ramesh2021zeroshottexttoimagegeneration}), have been able to produce convincing images of generic objects and scenes, but they struggle with representing a new, specific subject. To distill the information of a new subject into a diffusion model’s existing knowledge, one needs to update the vocabulary for a new token (\cite{gal2022imageworthwordpersonalizing}), and train the model again with the images of that subjet through expensively large computational resources, which is not feasible to do for a new object like a pet or a person. Naive solutions, such as fine-tuning all model parameters on a handful of subject images, often result in catastrophic forgetting, where the model’s broader knowledge is overwritten by the new concept. Few shot fine-tuning techniques including Dreambooth (\cite{ruiz2023dreamboothfinetuningtexttoimage}), Hyperdreambooth (\cite{ruiz2024hyperdreamboothhypernetworksfastpersonalization}) and Textual inversion (\cite{gal2022imageworthwordpersonalizing}) either have a high computational burden, or low image fidelity, limiting their adaptability. Furthermore, ensuring that the new subject fits seamlessly into a scene becomes difficult as the fine-tuning process significantly alters the model’s prior distribution.

In this work, we propose a two-stage pipeline that aims to preserve the original generative strengths of the latent diffusion-based (\cite{rombach2022highresolutionimagesynthesislatent}) image generation model Stable Diffusion XL (\cite{podell2023sdxlimprovinglatentdiffusion}) while introducing new subjects learned in a fast, lightweight manner. Our work include:

\begin{enumerate}[leftmargin=*]
    \item \textbf{Token selection:}
        We look for combination of letters in the CLIP (\cite{radford2021learningtransferablevisualmodels}) tokenizer that carry little to no prior association with any word. Prompting the model with these tokens yields varied or unrelated images, indicating that the model does not strongly link them to an existing concept.
    
    \item \textbf{LoRA-based fine-tuning:}
         We fine-tune only the attention layers of the Unet (\cite{ronneberger2015unetconvolutionalnetworksbiomedical}) of SDXL (\cite{podell2023sdxlimprovinglatentdiffusion}) with LoRA (\cite{hu2021loralowrankadaptationlarge}) on 4–5 subject images, such that the model overfits on the new object, and store the LoRA safetensors separately.
    
    \item \textbf{Two-Stage Generation:}
    \begin{enumerate}[leftmargin=*]
        \item \textbf{Stage 1:} Generate a generic image using the base (unmodified) SDXL by substituting the subject with its class label (e.g., “person”) in the prompt.
        \item \textbf{Stage 2:} Segmentation blurring using grounded SAM (\cite{ren2024groundedsamassemblingopenworld}) + Img2Img process to replace the segmented subject region with the newly learned concept using the LoRA safetensors, ensuring minimal interference with the overall scene.
    \end{enumerate}
    
    \item \textbf{Experimental Analysis:}
        We demonstrate that this approach efficiently combines the model’s compositional power with high-fidelity subject insertion, with the entire pipeline taking only about 7–8 minutes to complete on the SDXL model.
\end{enumerate}

Our empirical evaluation demonstrates that the proposed pipeline offers a fast, lightweight, and transferable solution to personalization in diffusion models. By isolating subject insertion from the broader generative distribution, we effectively mitigate challenges such as token selection, catastrophic forgetting, and excessive overfitting. We envision this approach as a stepping stone toward more robust, theoretically principled, and practically efficient methods for embedding novel concepts in deep generative models. This aligns with the workshop’s emphasis on bridging theoretical insights and practical efficacy, contributing to the advancement of DGMs in real-world applications.


\section{Related Work}

\subsection{Text-to-image generation}

Generative models have significantly evolved, with diffusion models (\citet{ho2020denoisingdiffusionprobabilisticmodels}, \cite{nichol2021improveddenoisingdiffusionprobabilistic} , \cite{rombach2022highresolutionimagesynthesislatent} , \cite{saharia2022paletteimagetoimagediffusionmodels} , \cite{saharia2021imagesuperresolutioniterativerefinement} , \cite{sohldickstein2015deepunsupervisedlearningusing} , \cite{song2022denoisingdiffusionimplicitmodels} , \cite{song2020generativemodelingestimatinggradients} , \cite{song2020improvedtechniquestrainingscorebased}), Generative Adversarial Networks (GANs)(\cite{brock2019largescalegantraining} , \cite{goodfellow2014generativeadversarialnetworks} , \cite{karras2021aliasfreegenerativeadversarialnetworks} , \cite{8953766} , \cite{9156570}), and transformer-based architectures driving advancements in text-to-image synthesis. Diffusion models, such as DALL·E 2 (\cite{ramesh2022hierarchicaltextconditionalimagegeneration}), Imagen (\cite{saharia2022photorealistictexttoimagediffusionmodels}), Stable Diffusion (\cite{rombach2022highresolutionimagesynthesislatent}), and MidJourney\footnote{https://www.midjourney.com} , have demonstrated remarkable capabilities by iteratively denoising a latent representation to refine random noise into coherent images. 

\subsection{Personalized Image Generation}

Several techniques have been developed to introduce personalization into diffusion-based image generation. DreamBooth (\cite{ruiz2023dreamboothfinetuningtexttoimage}) refines personalization by fine-tuning text-to-image models on a few user-provided images, allowing for the synthesis of subject-specific visuals while maintaining identity consistency. Textual Inversion (\cite{gal2022imageworthwordpersonalizing}) introduces a novel approach by learning new embeddings that encode personalized concepts within the latent space of a diffusion model, eliminating the need for full model fine-tuning. HyperDreamBooth (\cite{ruiz2024hyperdreamboothhypernetworksfastpersonalization}) improves these methods by incorporating hypernetworks, which improve adaptability and efficiency when fine-tuning diffusion models. Infusion (\cite{zeng2024infusionpreventingcustomizedtexttoimage}) prevents concept overfitting by categorizing it into concept-agnostic and concept-specific types but struggles to preserve image diversity. Cones (\cite{liu2023conesconceptneuronsdiffusion}) introduces concept neurons, focusing on interpretability over fine-tuning effectiveness, limiting its use for general customization. InstantBooth (\cite{shi2023instantboothpersonalizedtexttoimagegeneration}) eliminates test-time fine-tuning, it may struggle with capturing intricate identity details and handling complex concepts due to its reliance on global embeddings. Diffusion models have also been widely adopted for image editing, where methods like MyStyle (\cite{nitzan2022mystylepersonalizedgenerativeprior}) performed well in maintaining identity during image editing but needed extensive fine-tuning and computational power, restricting scalability. ForgeEdit (\cite{zhang2024forgedittextguidedimage}) enable fine-grained local modifications while maintaining global coherence.  InstructPix2Pi (\cite{brooks2023instructpix2pixlearningfollowimage}) extends this capability by integrating natural language instructions into the editing process, allowing users to specify precise changes using textual descriptions.

\subsection{Segmentation for Enhanced Control}

Segmentation models have played a crucial role in refining the control mechanisms for generative models. Grounded SAM (\cite{ren2024groundedsamassemblingopenworld}), which combines the capabilities of Grounding DINO (\cite{liu2024groundingdinomarryingdino}) and the Segment Anything Model (SAM) (\cite{kirillov2023segment}), allows for open-set object detection and segmentation guided by textual prompts. This integration enhances the applicability in tasks requiring region-specific transformations, such as targeted image editing, compositional generation, and object-focused synthesis.

\subsection{Efficient Adaptation using LoRA}

Fine-tuning large-scale diffusion models for personalized or domain-specific tasks is computationally expensive. Low-Rank Adaptation (LoRA) (\cite{hu2021loralowrankadaptationlarge}) addresses this challenge by decomposing weight updates into low-rank matrices, significantly reducing the number of trainable parameters while maintaining expressive power. Applied to diffusion models, LoRA enables efficient adaptation with limited computational resources, facilitating rapid personalization and domain adaptation without requiring full model retraining. This technique has proven particularly useful for adapting generative models to new tasks while preserving efficiency.
\begin{figure}[h]
\begin{center}
    \includegraphics[width=1\linewidth]{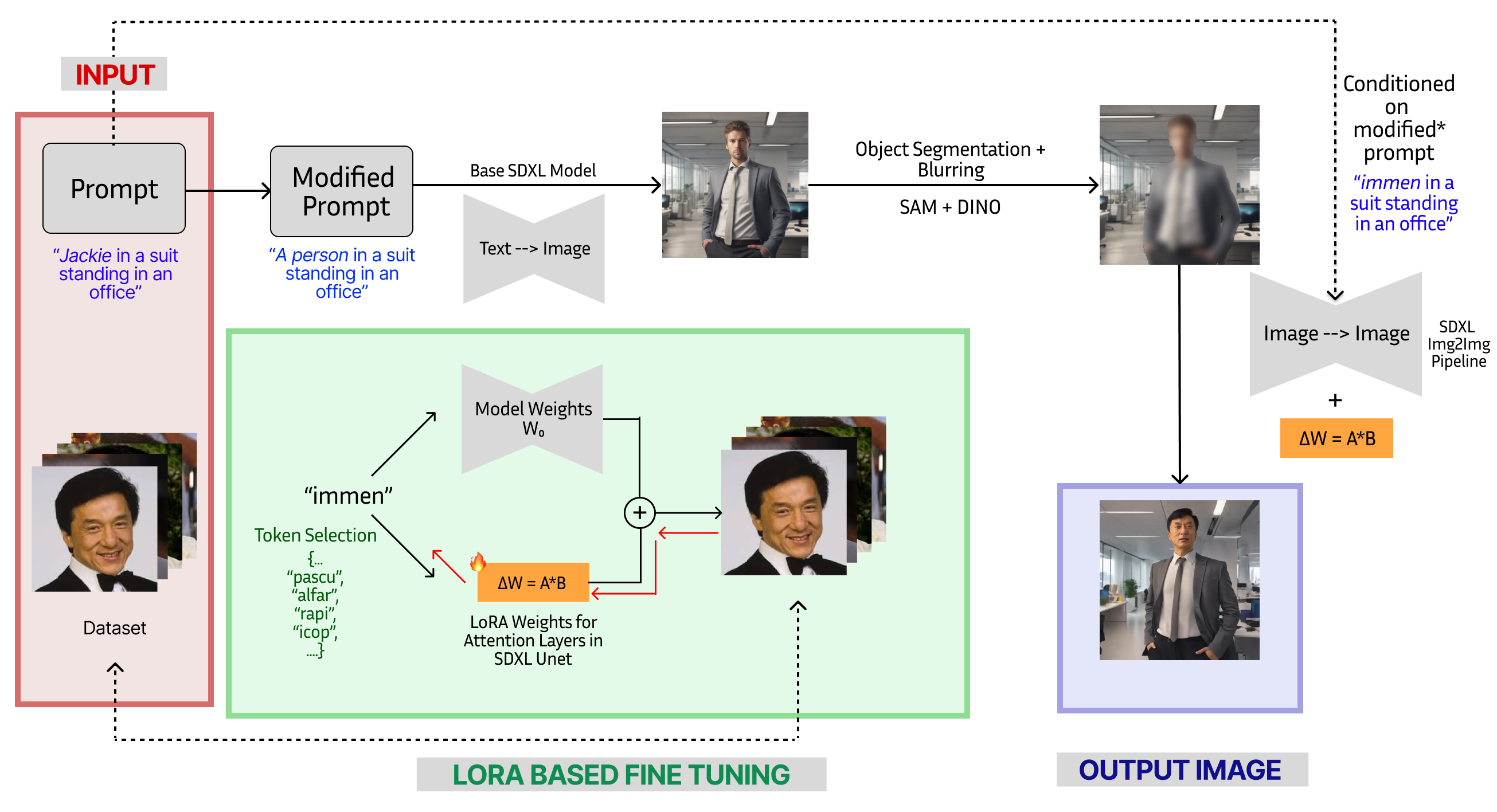} 
\end{center}
\caption{Overall Pipeline of Image-Alchemy}
\end{figure}

\section{Methodology}

\subsection{Token Selection}
\label{section:3.1}
In text-to-image diffusion models like SDXL, each prompt token is first mapped to a token embedding via CLIP’s text encoder. If an existing token already has a strong semantic association (e.g., “dog,” “cup,” or any common concept), using it to introduce a new subject can cause conflicts in the model’s learned representation. Consequently, we seek a “rare” token—one that the model does not strongly associate with any specific concept, and for which the CLIP tokenizer is not well trained on—so that the newly introduced subject can be learned without disrupting the model’s learned representation by a substantial margin.

\paragraph{Vocabulary Scan and Generation Test.}
We begin by scanning the CLIP tokenizer vocabulary for apparently gibberish tokens, typically short words (4–5 letters) that are unlikely to appear in conventional usage \footnote{For more details and our proposed tokens , see appendix}. For each candidate token, we perform multiple text-to-image generations using the unmodified SDXL under different random seeds. We then inspect the resulting images:
\begin{itemize}
    \item If the token consistently yields visually similar or thematically consistent outputs, it suggests the model associates it with an existing concept.
    \item If the token leads to highly varied or incoherent images, we infer that the model does not possess a strong prior for that token.
\end{itemize}
By comparing the generated images visually and through SSIM, we select the tokens that exhibit the greatest variability and least recognizable pattern.

Choosing a token with little to no prior allows the model to map our new subject’s features onto an essentially “blank slate.” When we train the model on a small set of subject images, the learned LoRA parameters effectively overwrite the token embedding and attention pathways tied to that placeholder. Because the base model had no strong associations with that token, the new concept is incorporated without disrupting other learned representations.

\subsection{LoRA based fine tuning}
\label{section:3.2}
Having selected a suitable placeholder token (\ref{section:3.1}), we next fine-tune the Stable Diffusion XL (SDXL) model on a small set of subject images. Instead of modifying all model parameters, as shown in Multi-Concept Customization of Text-to-Image Diffusion (\citep{kumari2023multiconceptcustomizationtexttoimagediffusion}) that during fine-tuning of diffusion models, the weight changes mostly occur in the cross-attention and self-attention layers, therefore we apply LoRA (Low-Rank Adaptation) to only a subset of the Unet of the Stable Diffusion-XL model weights. The Low Rank adaptation ensures prevention of catastrophic forgetting in the model's learned distribution. \footnote{For theoretical proof of the concept, see appendix}

\subsubsection{Intentional Overfitting}
Unlike conventional fine-tuning, our goal here is to ensure the model memorizes the subject’s unique features. We leverage the small training set (4–5 images) to push the LoRA parameters toward capturing fine details of the subject, resulting in overfitting. While overfitting is generally discouraged, we circumvent this problem by isolating the subject insertion to a later stage (\ref{section:3.4}).

\subsubsection{Storing LoRA Weights Separately}
The LoRA approach appends low-rank parameter matrices ($\Delta W$) to the original model weights. After training, we store these LoRA parameters (in .safetensors format) separately. This ensures non-destructive adaptation where the base model remains unchanged and flexible deployment such that a user can load or unload the specialized LoRA weights as and when required.

\subsection{Base image generation and segmentation}
\label{section:3.3}
The segmentation pipeline integrates Grounding DINO's zero-shot detection capabilities with SAM's prompt-driven segmentation in a two-step framework, resulting in mask generation. This methodology employs cross-modal alignment to achieve semantic segmentation and spatial refinement mechanisms to achieve pixel level boundaries in output masks.

\subsubsection{Detection Phase:}
Grounding DINO uses a Swin Transformer (\cite{liu2021swintransformerhierarchicalvision}) image encoder and a BERT-based (\cite{devlin2019bertpretrainingdeepbidirectional}) text encoder to project visual and textual features into a shared d-dimensional space ($d = 256$). Cross-modality fusion is achieved through attention mechanisms:
\begin{equation}
    A_{i,j} = \text{softmax}\left( \frac{\phi(f_i)^T \psi(e_j)}{\sqrt{d}} \right)
\end{equation}
where $\phi$ and $\psi$ are linear projections of image features ($f_i$) and text embeddings ($e_j$). The model outputs $N$ bounding boxes ($B \in \mathbb{R}^{N\times4}$) with confidence scores ($s \in [0,1]^N$), filtered using a threshold $\tau$ to remove low-confidence proposals.

\subsubsection{Segmentation Phase:}
SAM processes the detected boxes using a ViT-H/16 (\cite{dosovitskiy2021imageworth16x16words}) image encoder ($Z \in \mathbb{R}^{256\times H/16 \times W/16}$) and encodes box coordinates into sparse embeddings ($E_b \in \mathbb{R}^{N\times256}$). The mask decoder computes:
\begin{equation}
    M = \sigma\left( \text{MLP}\left( [Z \oplus (Z \odot E_b)] \right) \right)
\end{equation}
where $\oplus$ denotes concatenation and $\sigma$ is the sigmoid activation. The optimal mask is selected from multiple hypotheses using:
\begin{equation}
    M^* = \arg\max_{M_i} \left( \text{IoU}(M_i, B) - \lambda|M_i| \right)
\end{equation}
where $\lambda$ is a hyperparameter balancing mask precision and area coverage.

\subsection{Img2Img pipeline}
\label{section:3.4}
Once the base image is generated (\ref{section:3.3}) and the subject region is blurred, we load the LoRA-adapted SDXL model from Section \ref{section:3.2} and construct a prompt containing the placeholder token. For example, if the original user prompt was “A photo of Rahul sitting on a chair,” we replace “Rahul” with the placeholder token (e.g., “immen”) to obtain “A photo of immen sitting on a chair.” We then apply an Img2Img process: the blurred image serves as the initial input, and the prompt is applied as a condition, as the learned LoRA weights guide the diffusion model to fill in the blurred region with the newly introduced subject.

\subsubsection{Reverse Diffusion with Blurred Regions}
Latent diffusion models can be viewed as a stochastic differential equation (SDE):
\begin{equation}
    dx = -\nabla_x \log p_t(x) \, dt + \sqrt{\beta(t)} \, dW_t,
\end{equation}
where $x$ is the latent variable, and $-\nabla_x \log p_t(x)$ guides denoising. In blurred areas, gradient cues ($-\nabla_x \log p_t(x)$) are weaker because there is less high-frequency structure for the model to preserve. Consequently, the model relies more heavily on its learned priors—in our case, the subject-specific LoRA weights tied to the placeholder token. This mechanism enables accurate reconstruction of the newly introduced subject in regions designated for replacement.

\begin{figure}[h]
\begin{center}
    \includegraphics[width=1\linewidth]{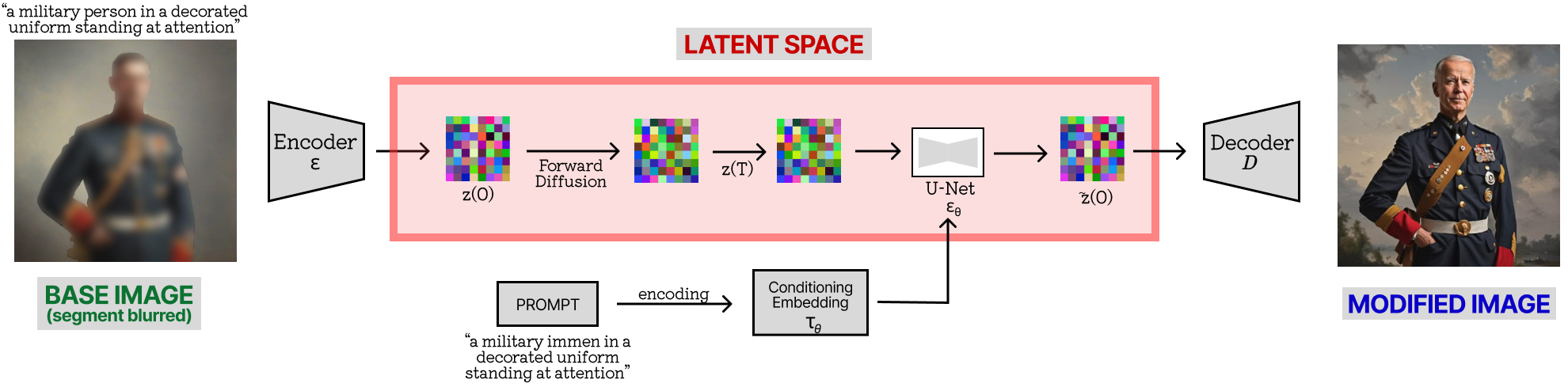} 
\end{center}
\caption{\textbf{Illustration of our Img2Img pipeline.} The base image is first segmented and blurred (left), then encoded into the latent space. The learned LoRA token (e.g., “immen”) is combined with the prompt and fed to the U-Net for iterative refinement, after which the decoder produces the final, personalized image (right).}
\label{figure:img-img-pipline}
\end{figure}

\subsubsection{Composition and style preservation}
Isolating the subject-insertion step after a generic image has already been produced removes the risk of catastrophic forgetting. Rather than relying on the fine-tuned model to generate the entire scene, we exploit SDXL’s original, unaltered capabilities for layout and background details. Consequently, the overfitted LoRA parameters only dominate where the subject must appear. This approach balances high-fidelity personalization with preserved compositional strength, which ensures integration of the new subject into the generated image effectively. Figure \ref{figure:img-img-pipline} illustrates how isolating subject insertion preserves composition while ensuring high-fidelity personalization.

\begin{figure}[h]
    \centering
    \includegraphics[width=0.5\linewidth]{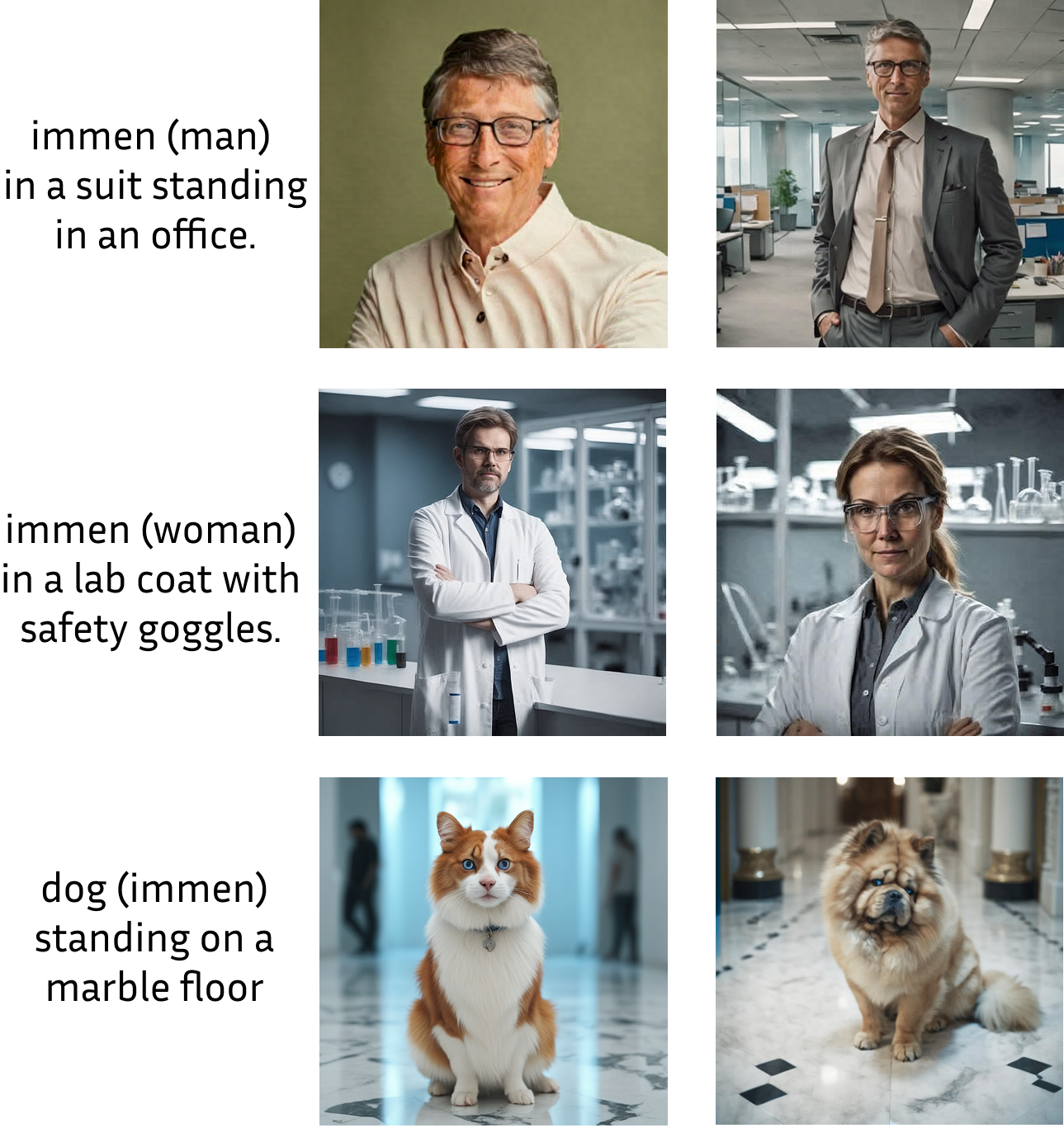}
    \caption{Column 1 has the prompts used to generate the images, Column 2 portrays the images generated using normal fine-tuning technique, the Column 3 contains the corresponding final output images of our pipeline.}
    \label{fig:cat_forget}
\end{figure}
\vspace{-10pt}
\section{Experiments}
\label{sec:experiments}

\begin{figure}[h]
    \centering
    \includegraphics[width=0.8\linewidth]{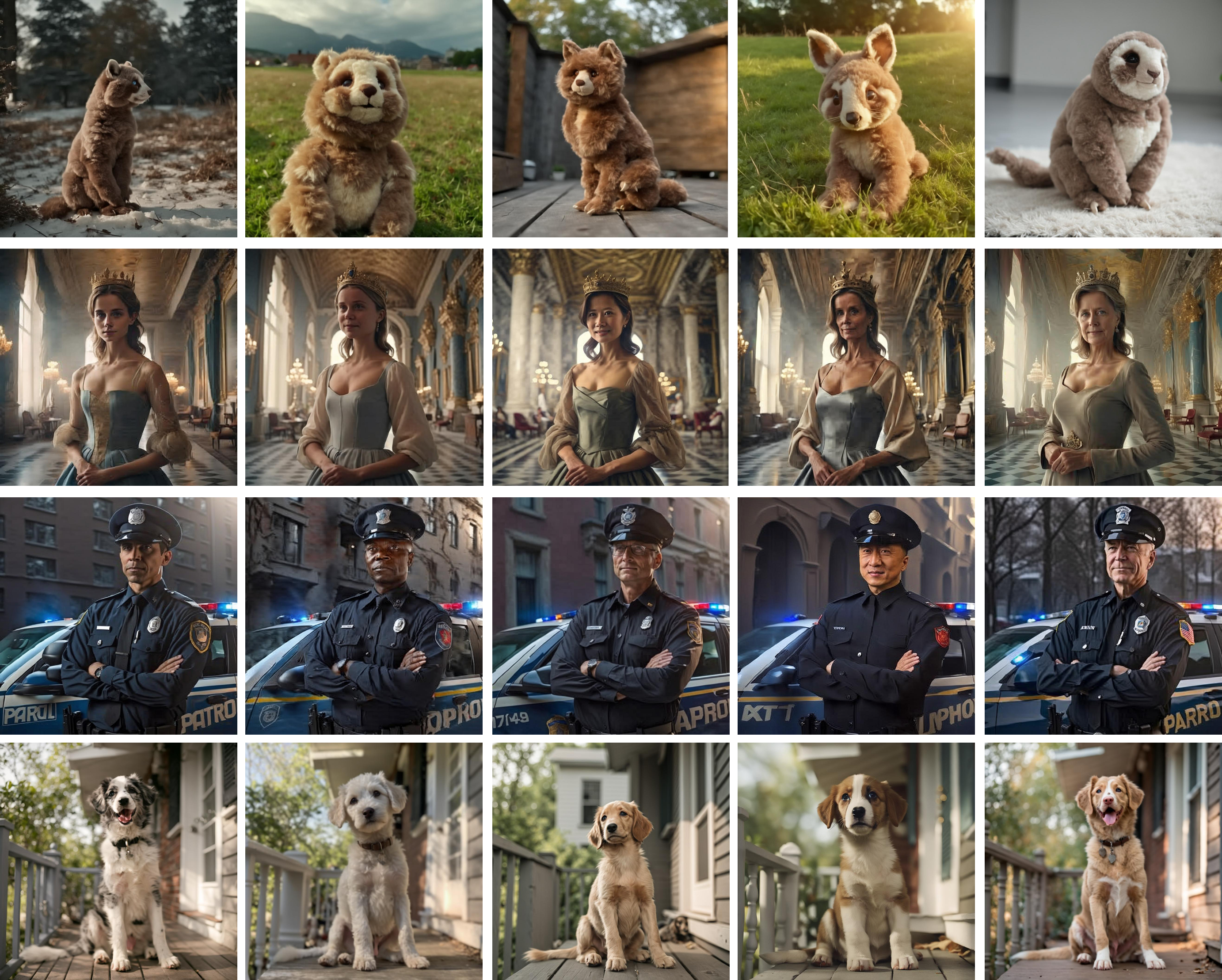} 
    \caption{\textbf{Seamless Subject Integration.} Row 1 displays a combination of bear and cat images, merging their characteristics while 
    preserving the overall quality and context.Row 2 presents images of various women, each integrated smoothly while keeping their unique identities.Row 3 features images of different men, with each image tailored to an individual while maintaining coherence.Row 4 showcases images of dogs, each distinctly personalized to a different dog, while ensuring a consistent style and quality..}
    \label{fig:diverse_subjects}
\end{figure}

\subsection{Direct Fine-Tuning and Catastrophic Forgetting}
\label{subsec:cat_forget}
We fine-tune SDXL on 4--5 reference images of a specific subject (e.g., ``Rahul'') using a standard approach similar to DreamBooth, without prior preservation loss.
 Figure \ref{fig:cat_forget} illustrates that the model loses its broader generative capabilities or produces images that do not align accurately with the given prompt. However, our model effectively preserves its generative diversity while ensuring that the generated images remain closely aligned with the input prompt, demonstrating a balanced trade-off between adaptability and fidelity.

\subsection{Subject Diversity}
\label{subsec:subject_diversity}
We started by evaluating the performance of our model in various subject categories \footnote{For the dataset details, see appendix}, including humans (both male and female) and animals (such as dogs and cats) to evaluate its adaptability and strength. To further examine its ability to generalize, after training the model on images of a specific subject, we provided a prompt for a different entity while using the learned token of the original subject. This led to impressive infusion effects, where characteristics of the trained subject were seamlessly integrated into the newly generated content as shown in Fig ~\ref{fig:diverse_subjects}.

\section{Results}
\label{sec:results}
\subsection{Embedding-Based Similarities}
\label{subsec:embedding_results}

We compute embeddings for both reference images and generated outputs using \textbf{CLIP} (CLIP-I) (\cite{radford2021learningtransferablevisualmodels}) and \textbf{DINO} (\cite{caron2021emergingpropertiesselfsupervisedvision}), by measuring cosine similarity in the subject region. Higher values indicate better subject preservation.

These metrics are used to evaluate how well the generated outputs preserve the subject from the reference images. \textbf{CLIP-I} measures cosine similarity between the embeddings of reference and generated images, but it is not considered a reliable metric for evaluating frameworks like ours because it fails to recognize rare or unique tokens (e.g., "mccre wearing a leather jacket and jeans") that hold no meaning in its text encoder, leading to arbitrary similarity scores. \textbf{DINO} evaluates subject preservation by computing cosine similarity. DINO generally performs well in capturing visual features and is less reliant on text-based understanding compared to CLIP. \textbf{CLIP-T} measures cosine similarity between textual descriptions of the generated and reference images. It measures how well the generated image aligns with textual prompts.

\begin{table}[h]
\caption{CLIP and DINO embedding similarities (cosine) between final outputs and reference images. Higher is better.}
\label{table:clip_dino}
\begin{center}
\begin{tabular}{lccc}
\multicolumn{1}{c}{\bf METHOD} & \multicolumn{1}{c}{\bf DINO} & \multicolumn{1}{c}{\bf CLIP-I} & \multicolumn{1}{c}{\bf CLIP-T} \\
\hline \\
Real & 0.834 & 0.763  & NA  \\
Dreambooth(Stable Diffusion) & 0.668 & \textbf{0.803} & 0.305\\
Textual Inversion(Stable Diffusion) & 0.569 & 0.780 & 0.255\\   
Custom Diffusion & 0.643 & 0.798 & 0.256 \\
Subject Diffusion & 0.711 & 0.787 & 0.293 \\
Ours (Two-Stage) & \textbf{0.789} & 0.557 & \textbf{0.334} \\
\end{tabular}
\end{center}
\end{table}

\subsection{Image Quality Metrics}
\label{subsec:quality_metrics}
We assess image fidelity using metrics such as 
\textbf{NIQE} (\cite{6353522}) which quantifies naturalness by measuring statistical distances between multivariate Gaussian models of test image features and those of pristine natural images, using MSCN coefficients from local patches, 
\textbf{BRISQUE} (\cite{6272356}) evaluates distortions through SVR-based analysis of locally normalized luminance statistics' deviation from natural image models in the spatial domain and, 
\textbf{MANIQA} (\cite{yang2022maniqa}) a transformer-based IQA using multi-head self-attention to learn hierarchical quality features through spatial and channel attention weighting of image patches, where applicable. 

\begin{table}[h]
\caption{Blind Image quality metrics}
\label{table:image_quality}
\begin{center}
\begin{tabular}{lccc}
\multicolumn{1}{c}{\bf METHOD}  & \multicolumn{1}{c}{\bf NIQE $\uparrow$} & \multicolumn{1}{c}{\bf MANIQA $\uparrow$} & \multicolumn{1}{c}{\bf BRISQUE $\downarrow$ } \\
\hline \\
Real Images & 7.837 & 0.5264 & 23.644 \\ 
Unmodified SDXL & 5.681 & 0.3930 & 45.047 \\
Ours (Two-Stage) & 5.727 & 0.3571 & 46.302 \\
\end{tabular}
\end{center}
\end{table}
Table~\ref{table:image_quality} suggests that our two-stage pipeline successfully personalizes the subject while minimally impacting the broader scene quality.The difference between SDXL’s baseline generation and our personalized outputs remains small, suggesting minimal degradation in overall fidelity. As expected, real images achieve the best metric scores.
\vspace{-10pt}
\section{Conclusion}
\vspace{-10pt}
In conclusion, our framework for personalized text-to-image generation addresses key challenges in adapting diffusion models to novel subjects while maintaining their broader generative capabilities. By leveraging LoRA-based fine-tuning on SDXL's attention layers and employing a segmentation-driven Img2Img process, the method effectively isolates subject personalization from scene composition. This approach mitigates issues such as catastrophic forgetting and token selection conflicts, while significantly reducing computational costs.
The experimental results demonstrate that this pipeline achieves high-fidelity subject integration with minimal disruption to the original model's versatility. The modular nature of LoRA weights ensures flexible deployment without altering the base model, making this solution practical for real-world applications.
This work represents a significant step forward in personalized image synthesis, offering a lightweight and efficient framework that bridges theoretical advancements with practical utility. Future research could explore extending this approach to multiple and more complex subjects,further enhancing its adaptability and robustness in diverse use cases.

\bibliography{iclr2025_delta}
\bibliographystyle{iclr2025_delta}

\appendix
\section{Appendix}
\subsection{Blurring Techniques}
We integrated selective blurring of the main object in the base image before passing it to the img-to-img SDXL pipeline. This step was designed to enhance the quality of edits by focusing modifications in proximity to the primary object of interest. Two distinct blurring techniques were evaluated for this purpose:
\begin{itemize}
    \item \textbf{Gaussian Blur:}  
    In this method, we applied a Gaussian blur to the region identified by the mask generated by the Grounded Segment Anything Model (SAM). The blur was applied uniformly across the masked region with kernel\_size=151, $\sigma = 100$ for the Gaussian kernel.
    \[
        G(x, y) = \frac{1}{2\pi\sigma^2} \exp\left(-\frac{x^2 + y^2}{2\sigma^2}\right)
    \]
    
    \item \textbf{Exponential Decay Gaussian Blur:}  
    This technique introduced a more nuanced approach by incorporating an exponential decay factor into the Gaussian blur. The degree of blurring decreased exponentially with increasing distance from the main object, as determined by the nearest zero value in the mask generated by Grounded SAM. (kernel\_size=151, $\sigma = 100$, $\lambda = 5$)
    \[
        G'(x, y) = G(x, y) \cdot e^{-\lambda d}
    \]
\end{itemize}
The experimental results indicate that the second technique consistently produces superior results. The key reason behind this improvement is that an exponential decay in blurring ensures a smoother transition between the blurred and unblurred regions. Unlike the fixed Gaussian blur, which applies a uniform blur within the segmented area and often leads to unnatural transitions, the decay-based approach maintains spatial coherence by maintaining the details in the background of the object and smoother transitions in the vicinity of the object.

\subsection{Tokens}
We found that the tokens used by DreamBooth for image personalization show strong biases in the CLIP text encoder of the Stable Diffusion XL (SDXL) model. For example, the token "sks" often led to the generation of a firearm when used for subject personalization. This indicates that the token had a strong existing meaning in the model, which can harm subject adaptation.

To address this problem, we propose a new set of tokens for subject personalization in SDXL. These tokens are chosen for their limited use in the text encoder, ensuring that the model has little to no prior understanding of them. This allows for more accurate and unbiased subject learning, resulting in better image personalization without interference from existing associations. Using these tokens encode and represent the personalized subject more effectively, producing more reliable and controllable results. Such tokens are listed below:\[
\begin{array}{cccc}
\text{immen} & \text{pasqu} & \text{iklan} & \text{rapi}  \\
\text{bhar}  & \text{ellu}  & \text{ffin}  & \text{icop}  \\
\text{aben}  & \text{mmor}  & \text{psal}  & \text{phyl}  \\
\text{rrrr}  & \text{wozni} & \text{geaux} & \text{koval} \\
\text{ayles} & \text{mccre} & \text{fortn} & \text{prote} \\
\text{pascu} & \text{lisam} & \text{percu} & \text{alfar} \\
\text{insom} & \text{offro} & \text{syour} & \text{redon} \\
\end{array}
\]

\subsection{Dataset}
\begin{figure}[h]
\begin{center}
    \includegraphics[width=0.7\linewidth]{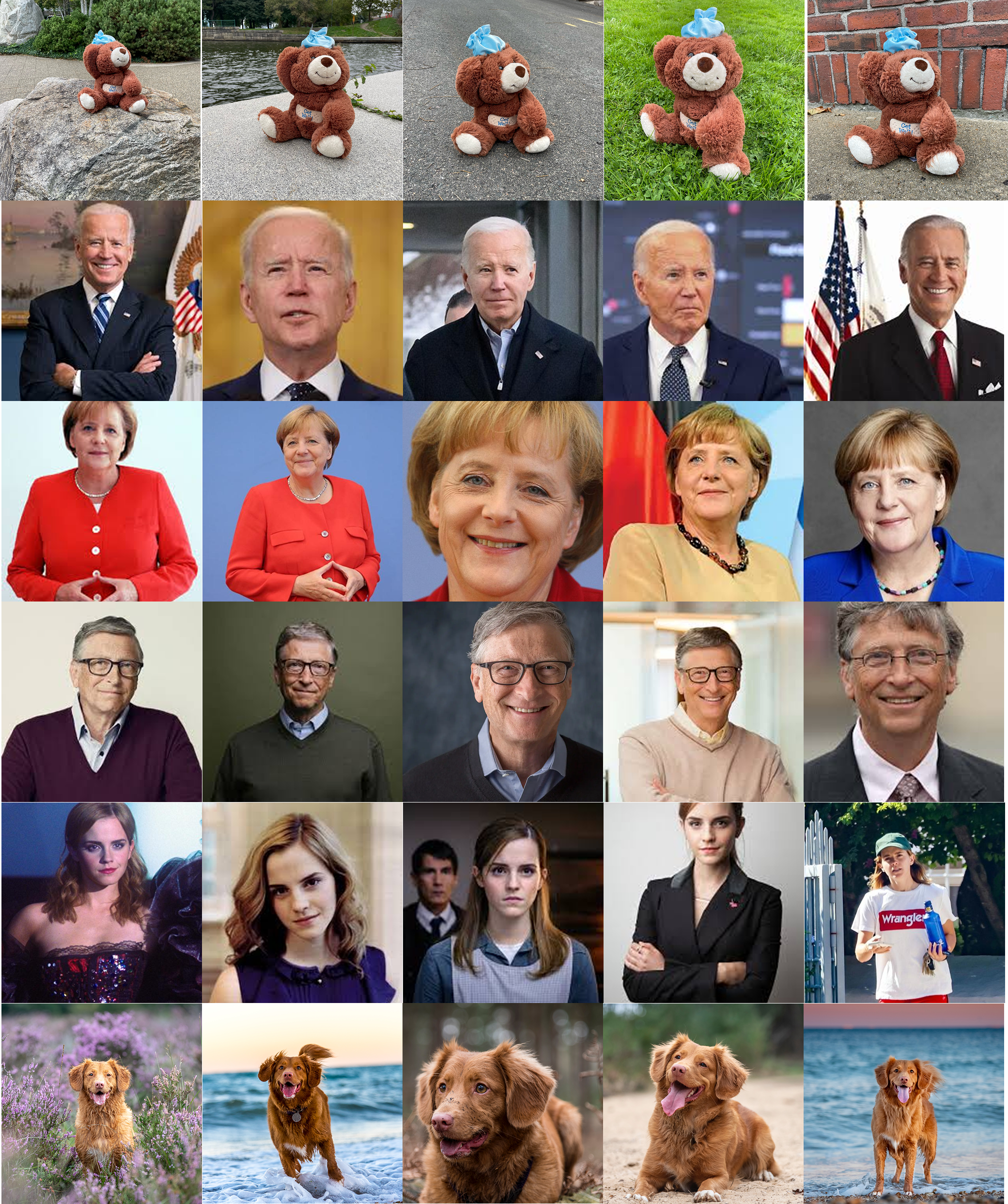} 
\end{center}
\caption{An instance from the datasets used to fine tune Stable Diffusion XL model.}
\label{fig:dataset_instance}
\end{figure}

We used data from various sources to strengthen our fine-tuning at different stages of our process. We selected images that are commercially licensed and copyright-free to meet legal and ethical standards and for object-specific fine-tuning, we included the dataset from the DreamBooth paper. Figure \ref{fig:dataset_instance} illustrates a sample dataset instance used for metric evaluation and pipeline validation.

\subsection{Catastrophic Forgetting Mitigation through Low Rank Adaptation}
\label{app:theory_catastrophic_forgetting}

In our setup, the base Stable Diffusion XL model is parameterized by $\theta \in \mathbb{R}^D$, and we learn a small set of LoRA parameters $\Delta \theta \in \mathbb{R}^{d}$ (where $d \ll D$) that modify only the attention layers. Let $p_\theta(\mathbf{x} \mid \mathbf{y})$ denote the model’s distribution over images $\mathbf{x}$ given prompt $\mathbf{y}$. After LoRA adaptation, the parameters become $(\theta, \Delta \theta)$, inducing a (slightly) modified distribution $p_{\theta + \Delta \theta}(\mathbf{x} \mid \mathbf{y})$.

We introduce the LoRA update as:
\begin{equation}
    \label{eq:lora_param}
    W_{\text{LoRA}} = W + \alpha \cdot U V^\top,
\end{equation}
where $W \in \mathbb{R}^{d_1 \times d_2}$ is a pre-trained attention weight matrix, $U \in \mathbb{R}^{d_1 \times r}$ and $V \in \mathbb{R}^{r \times d_2}$ define a rank-$r$ update (with $r \ll \min(d_1, d_2)$), and $\alpha$ is a scaling constant ($\alpha$ = 1 in our case). Only $U, V$ are optimized, leaving $W$ fixed. Consequently, the effective update $\Delta W = \alpha \cdot U V^\top$ occupies a low-dimensional subspace.
\paragraph{Bound on Distributional Shift.}
We can characterize the distributional shift via a divergence measure, e.g. the Kullback-Leibler (KL) divergence:
\begin{equation}
    \label{eq:distribution_shift}
    D_\mathrm{KL}\!\bigl(p_{\theta + \Delta \theta}(\mathbf{x} \mid \mathbf{y}) \,\big\|\,
    p_{\theta}(\mathbf{x} \mid \mathbf{y})\bigr).
\end{equation}
Under assumptions of Lipschitz continuity in $\theta$-space, we can bound \eqref{eq:distribution_shift} by a function of $\|\Delta \theta\|$ (or $\|\Delta W\|$). Since $\Delta W$ is rank-$r$ , we have:
\[
    \|\Delta W\|_F 
    \;=\;
    \|\alpha \cdot U V^\top\|_F 
    \;\le\;
    \alpha \,\|U\|_F \,\|V^\top\|_F
    \;=\;
    \alpha\,\|U\|_F \,\|V\|_F,
\]
which is significantly smaller than a full $d_1 \times d_2$ dimensional update if $r$ is controlled. Thus,
\[
    D_\mathrm{KL}\!\bigl(p_{\theta + \Delta \theta}, p_{\theta}\bigr)
    \;\le\;
    \kappa \cdot \|\Delta W\|_F,
\]
for some Lipschitz constant $\kappa > 0$. Therefore, if $\|\Delta W\|_F$ is kept small, the change in the global distribution $p_\theta(\mathbf{x}\mid \mathbf{y})$ remains bounded, preserving much of the original knowledge.

Basically, \emph{distributional shift} from $p_\theta$ to $p_{\theta + \Delta \theta}$ depends on $\|\Delta W\|_F$, which in turn is bounded by $\alpha\, \|U\|_F \,\|V\|_F$. Thus, a small-rank update constrains how \emph{strongly} the new subject can overwrite existing representations, mitigating forgetting.

\paragraph{When Might LoRA Fail?}
If the new subject is extremely complex or demands rewriting fundamental visual features, a low-rank subspace might be insufficient, leading to partial adaptation. In that case, there is incomplete personalization or minor forgetting for other concepts. However, for typical tasks (like injecting a single person or object), this rank-limited approach is both sufficiently expressive and far safer than full-model fine-tuning.

\end{document}